%% file: main.tex
\documentclass[10pt,twocolumn,letterpaper]{article}
\usepackage[pagenumbers]{cvpr}

\usepackage{graphicx}
\usepackage{amsmath,amssymb}
\usepackage{booktabs}
\usepackage{multirow}
\usepackage{xcolor}
\usepackage{xspace}
\usepackage{enumitem}
\usepackage{caption}

\newcommand{\spotdiff}{SpotDiff\xspace}

\definecolor{cvprblue}{rgb}{0.21,0.49,0.74}
\usepackage[pagebackref,breaklinks,colorlinks,allcolors=cvprblue]{hyperref}

\hypersetup{
  pdftitle={SpotDiff: A Plug-and-Play Framework for Zero-Shot Change Detection in Remote Sensing Imagery},
  pdfauthor={Anit Upadhyaya, Ata Akbari Asanjan, Chaitanya Jalluri, Subash Yeggina, Jeremy Kravitz},
  pdfsubject={Zero-shot and open-vocabulary change detection in remote sensing},
  pdfkeywords={change detection, zero-shot change detection, open-vocabulary change detection, OVCD, training-free, remote sensing, Earth observation, vision foundation models, DINOv3, SAM 3, feature upsampling, AnyUp, building change detection, semantic change detection}
}

\def\confName{arXiv}
\def\confYear{2026}

\title{SpotDiff: A Plug-and-Play Framework for Zero-Shot\\Change Detection in Remote Sensing Imagery}

\author{%
Anit Upadhyaya$^{1}$\enspace Ata Akbari Asanjan$^{2}$\enspace Chaitanya Jalluri$^{1}$\enspace Subash Yeggina$^{1}$\enspace Jeremy Kravitz$^{2}$\\[4pt]
$^{1}$Pixxel Space India Pvt Ltd, Bengaluru, India\\
$^{2}$Pixxel Space Technologies, El Segundo, United States
}

\begin{document}
\maketitle

\input{sections/0_abstract}
\input{sections/1_intro}
\input{sections/2_related_work}
\input{sections/3_method}
\input{sections/4_experiments}
\input{sections/5_conclusion}

{
\small
\bibliographystyle{ieeenat_fullname}
\bibliography{refs}
}

\end{document}

%% file: sections/0_abstract.tex
\begin{abstract}
Remote sensing change detection usually relies on large collections of temporally aligned, pixel-level annotations, which limits its scalability across regions, imaging platforms and application domains.
Recent zero-shot approaches reduce the need for supervision, but they often localize fine-grained changes poorly because their features are coarse or inconsistent across time.
We present \spotdiff, a training-free, plug-and-play framework for zero-shot change detection built by composing complementary vision foundation models.
\spotdiff uses DINOv3 for semantically robust features, AnyUp to reconstruct them at pixel resolution, and SAM~3 for text-guided region-of-interest proposals, which an inference-time multi-scale strategy (SWIMS) makes reliable on small objects.
Change is then measured by comparing the reconstructed features of the two dates inside the proposed regions, with no learnable components, fine-tuning or change labels.
To our knowledge, \spotdiff is the first change detection method to use a universal feature upsampler, which alone adds 13.8 F1 over bilinear upsampling of the same features. On LEVIR-CD, \spotdiff reaches 82.46 F1, 2.06 points above the strongest prior training-free method, and on the multi-class SECOND benchmark it improves building change detection by 10.3 IoU and the six-class average by 2.2 IoU.
Qualitative results on PlanetScope (3\,m) and Sentinel-2 (10\,m) imagery show that the same pipeline transfers across resolutions without modification.
Our results indicate that composing modern foundation models is an effective route to scalable, label-free change detection.
\end{abstract}

\noindent\textbf{Keywords:} change detection, zero-shot change detection, open-vocabulary change detection (OVCD), training-free, remote sensing, Earth observation, vision foundation models, DINOv3, SAM~3, feature upsampling, building change detection, semantic change detection.

%% file: sections/1_intro.tex
\section{Introduction}
\label{sec:intro}

Change detection, the task of identifying meaningful differences between images of the same area acquired at different times, underpins urban expansion monitoring, disaster assessment, infrastructure development, environmental monitoring and precision agriculture~\cite{chen2020levir,daudt2018fully}.
Most state-of-the-art methods are trained on large collections of temporally aligned, pixel-level annotations~\cite{chen2020levir,fang2021snunet,bandara2022changeformer}.
Such annotations are expensive to produce and rarely available across regions, imaging platforms and application domains, and supervised models often generalize poorly outside the distribution they were trained on.

Vision foundation models have changed this picture.
Self-supervised vision transformers such as DINOv3~\cite{simeoni2025dinov3} learn semantically rich, transferable representations, and segmentation foundation models~\cite{kirillov2023sam,ravi2024sam2,carion2025sam3} localize objects zero-shot.
This raises a direct question for Earth observation:
\begin{quote}
\emph{Can modern vision foundation models already perform accurate zero-shot change detection in remote sensing, without any task-specific training?}
\end{quote}

The answer is not obvious.
Foundation models are not designed for bi-temporal analysis.
Vision transformer features are produced at patch resolution (one vector per 16$\times$16 pixels for DINOv3), too coarse to delineate small structures such as buildings.
Semantic similarity alone produces noisy change maps, because illumination, season and acquisition geometry also change between dates.
And nothing in these models restricts the analysis to the objects a user cares about.

We present \spotdiff, a training-free, plug-and-play framework that answers this question by composition (\cref{fig:arch}).
\spotdiff combines DINOv3 for semantic features, AnyUp~\cite{wimmer2025anyup} to reconstruct them at pixel resolution, and SAM~3~\cite{carion2025sam3} to propose regions of the queried category from a text prompt.
To make SAM~3 reliable on small objects in large tiles, we introduce SWIMS (sliding-window inference with multiple scales), which fuses proposals from several input scales.
Change is declared where the reconstructed features of the two dates disagree inside the proposed regions.
The framework has no learnable components, needs no temporal supervision, and lets each component be replaced as stronger models appear.

The key observation behind \spotdiff is that modern self-supervised features are stable enough across acquisition dates that, once restored to pixel resolution and restricted to semantic region priors, plain cosine similarity localizes meaningful change.
In other words, strong zero-shot change detection can emerge from a careful composition of existing models, without a dedicated change detection architecture.

We evaluate \spotdiff on LEVIR-CD, WHU-CD, S2Looking and the multi-class SECOND benchmark, and show qualitative results on PlanetScope (3\,m) and Sentinel-2 (10\,m) imagery.
Among training-free methods published before April 2026, \spotdiff sets a new state of the art on LEVIR-CD (82.46 F1), WHU-CD and S2Looking, and on building change and the class average of SECOND.

\noindent\textbf{Contributions.}
\begin{itemize}[leftmargin=*,itemsep=1pt,topsep=2pt]
  \item We present \spotdiff, a training-free, plug-and-play framework for zero-shot remote sensing change detection that composes a self-supervised encoder, a universal feature upsampler and a promptable segmentation model. To our knowledge, it is the first change detection method to use a universal feature upsampler, and we show that pixel-resolution features are the single most important ingredient after multi-scale proposals.
  \item We show that comparing pixel-resolution self-supervised features inside text-guided region priors localizes change accurately without task-specific optimization, and we introduce SWIMS, a multi-scale inference strategy that makes the region priors reliable for small objects.
  \item We evaluate \spotdiff on four public benchmarks and two additional sensors, with ablations of the upsampler, SWIMS, the similarity threshold and the text prompt, establishing a strong training-free baseline.
\end{itemize}

%% file: sections/2_related_work.tex
\section{Related Work}
\label{sec:related}

\paragraph{Supervised change detection.}
Early deep methods adapted encoder-decoder networks to bi-temporal inputs; FC-EF, FC-Siam-conc and FC-Siam-diff~\cite{daudt2018fully} introduced siamese fusion variants.
Later work added spatial-temporal attention (STANet~\cite{chen2020levir}), dense connections with deep supervision (SNUNet~\cite{fang2021snunet}), transformers (BIT~\cite{chen2021bit}, ChangeFormer~\cite{bandara2022changeformer}), lightweight designs (TinyCD~\cite{codegoni2023tinycd}), state-space models (RS-Mamba~\cite{zhao2024rsmamba}) and adapted SAM encoders (SAM-CD~\cite{ding2024samcd}).
ChangeCLIP~\cite{dong2024changeclip} brings vision-language representations into supervised change detection.
These methods are accurate in-domain but require pixel-level annotations aligned across time.

\paragraph{Foundation models for remote sensing.}
DINO~\cite{caron2021dino} and DINOv2~\cite{oquab2023dinov2} showed that self-supervised vision transformers learn transferable features; DINOv3~\cite{simeoni2025dinov3} scales this recipe and improves dense feature quality.
SAM~\cite{kirillov2023sam} introduced promptable zero-shot segmentation, SAM~2~\cite{ravi2024sam2} extended it to video, and SAM~3~\cite{carion2025sam3} segments all instances of a concept given a short noun phrase.
Remote sensing adaptations such as RSPrompter~\cite{chen2024rsprompter} and GeoSAM~\cite{ibn2023geosam} fine-tune these models for specific tasks, which departs from the zero-shot setting.

\paragraph{Zero-shot and open-vocabulary change detection.}
Classical unsupervised methods rely on change vector analysis, clustering or deep feature differencing~\cite{bovolo2006cva,celik2009pcakm,saha2019dcva,du2019dsfa,tang2021gmcd}.
AnyChange~\cite{zheng2024anychange} detects class-agnostic changes training-free by bitemporal latent matching in the SAM feature space, and SCM~\cite{tan2024scm} combines SAM with CLIP for unsupervised building change.
BiSAM-CD~\cite{qin2025bisamcd} propagates masks with the temporal memory of SAM~2.
DynamicEarth~\cite{li2025dynamicearth} formalizes open-vocabulary change detection (OVCD) and benchmarks training-free pipelines composed of mask proposers, comparators and identifiers.
OmniOVCD~\cite{zhang2026omniovcd} streamlines OVCD with SAM~3 by fusing its semantic, instance and presence outputs and matching instances across dates. AdaptOVCD~\cite{dou2026adaptovcd} fuses SAM-HQ masks, interpolated DINOv3 features and a remote sensing CLIP with adaptive thresholds, and OpenDPR~\cite{guo2026opendpr} retrieves class prototypes from diffusion-synthesized support images and compares DINOv2 features.
\spotdiff shares the training-free, compositional spirit of these works, but decides change by comparing pixel-resolution self-supervised features rather than masks or SAM latents, and uses the segmentation model only to restrict where change is measured.

\paragraph{Feature upsampling.}
Vision transformers output features at patch resolution.
FeatUp~\cite{fu2024featup} learns upsampling through multi-view consistency but must be trained per encoder.
AnyUp~\cite{wimmer2025anyup} is encoder-agnostic: trained once, it upsamples features from encoders it has never seen, which lets \spotdiff change its backbone without retraining anything. Change detection pipelines that compare foundation-model features have so far interpolated patch features to image resolution; to our knowledge, \spotdiff is the first to use a universal feature upsampler for change detection.

%% file: sections/3_method.tex
\section{Method}
\label{sec:method}

\begin{figure*}[t]
  \centering
  \includegraphics[width=\linewidth]{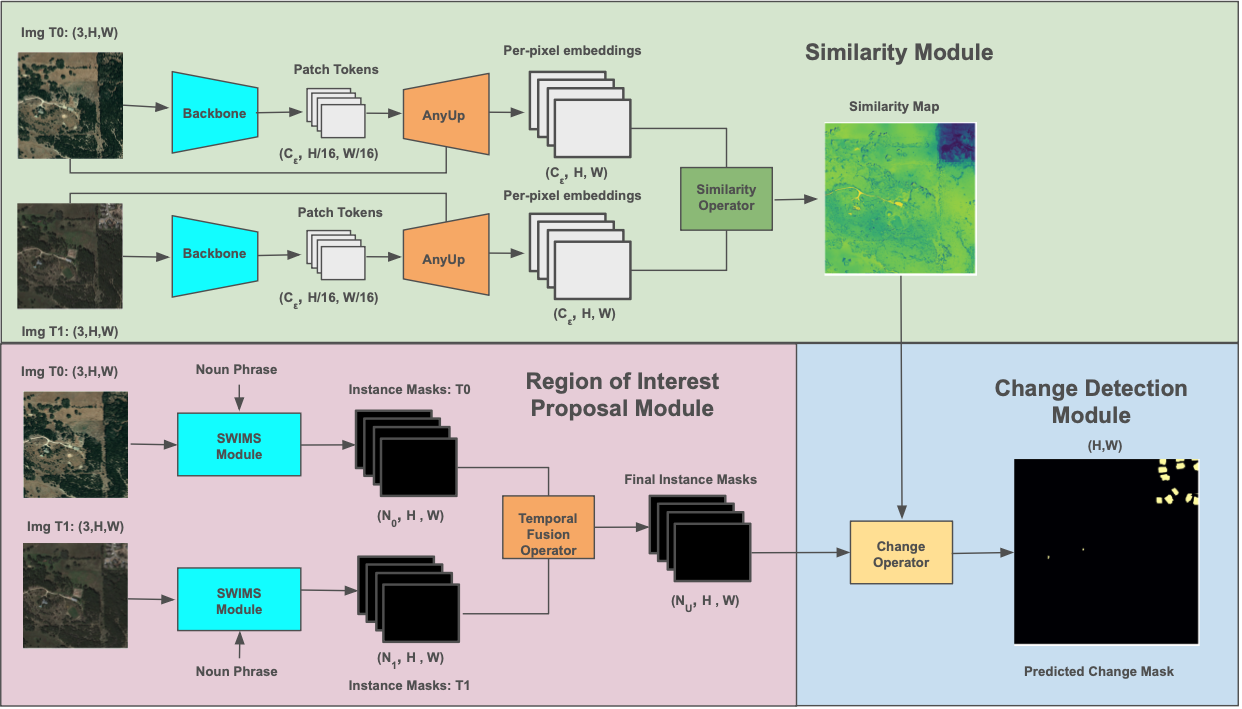}
  \caption{\textbf{Overview of \spotdiff.} The Similarity Module encodes both images with a frozen vision foundation model, reconstructs pixel-resolution features with a universal upsampler and computes their per-pixel cosine similarity. The Region of Interest Proposal Module segments instances of the prompted category in each image with multi-scale inference (SWIMS) and fuses them across time. The Change Detection Module marks a candidate instance as changed when its mean similarity falls below a threshold. No component is trained.}
  \label{fig:arch}
\end{figure*}

\spotdiff separates \emph{what changed} from \emph{where to look}.
As shown in \cref{fig:arch}, it has three components: a Similarity Module that measures semantic consistency across time at pixel resolution, a Region of Interest (RoI) Proposal Module that finds instances of the queried category, and a Change Detection Module that combines the two.
Every component is a frozen, off-the-shelf model or a parameter-free operation, so each can be replaced independently.

\subsection{Problem formulation}
Given two co-registered images $\mathbf{I}_0, \mathbf{I}_1 \in \mathbb{R}^{3 \times H \times W}$ acquired at different times, and an optional text prompt $\mathbf{p}$ naming the category of interest (\eg, \emph{building}), the goal is a binary change mask $\mathbf{Y} \in \{0,1\}^{H \times W}$ with $Y(i,j)=1$ where pixel $(i,j)$ belongs to a change of that category.
\spotdiff assumes no temporal annotations, no fine-tuning and no task-specific optimization.
It can be written as the composition
\begin{equation}
  \mathbf{Y} = \mathcal{C}\big(\mathcal{S}(\mathbf{I}_0,\mathbf{I}_1),\; \mathcal{R}(\mathbf{I}_0,\mathbf{I}_1,\mathbf{p})\big),
  \label{eq:overview}
\end{equation}
where $\mathcal{S}$, $\mathcal{R}$ and $\mathcal{C}$ are the Similarity, RoI Proposal and Change Detection modules.

\subsection{Similarity Module}
Vision transformers produce features at patch resolution, and comparing them directly yields blurry similarity maps that miss small changes.
\spotdiff therefore reconstructs pixel-level features before comparing the two dates.
A frozen encoder $f_\theta$ (DINOv3~\cite{simeoni2025dinov3}) maps each image to patch features
\begin{equation}
  \mathbf{Z}_t = f_\theta(\mathbf{I}_t) \in \mathbb{R}^{C \times \frac{H}{16} \times \frac{W}{16}}, \quad t \in \{0,1\},
\end{equation}
and a universal upsampler $g_\phi$ (AnyUp~\cite{wimmer2025anyup}), guided by the image, restores full resolution:
\begin{equation}
  \mathbf{E}_t = g_\phi(\mathbf{Z}_t, \mathbf{I}_t) \in \mathbb{R}^{C \times H \times W}.
\end{equation}
The similarity map is the per-pixel cosine similarity of the two dates,
\begin{equation}
  S(i,j) = \frac{\mathbf{E}_0(i,j)^\top \mathbf{E}_1(i,j)}{\lVert \mathbf{E}_0(i,j) \rVert \, \lVert \mathbf{E}_1(i,j) \rVert}.
\end{equation}
Because the comparison happens in a semantic feature space rather than on pixel values, it is far less sensitive to illumination and radiometric differences than image differencing, while still responding to a change of content.

\subsection{Region of Interest Proposal Module}
A promptable segmentation model $h$ (SAM~3~\cite{carion2025sam3}) returns the instances of the prompted category in each image,
\begin{equation}
  \mathbf{M}_t = h(\mathbf{I}_t, \mathbf{p}) = \{m_1, \ldots, m_{N_t}\}, \quad m_k \in \{0,1\}^{H \times W}.
\end{equation}

\paragraph{SWIMS.}
On large tiles, a single forward pass misses many small or partially visible instances.
SWIMS (sliding-window inference with multiple scales) upsamples the image by a factor $s$, runs $h$ on sliding windows at the model's native input size, and maps the predictions back:
\begin{equation}
  \mathbf{M}_t = \bigcup_{s \in \mathcal{K}} \mathrm{Down}_{1/s}\big(h(\mathrm{Up}_s(\mathbf{I}_t), \mathbf{p})\big),
\end{equation}
with bicubic $\mathrm{Up}_s$.
We use $\mathcal{K} = \{1, 2\}$: the $2\times$ scale recovers small objects, while adding $4\times$ introduces false positives from over-magnified, uniform regions (\cref{sec:ablation}).

\paragraph{Temporal fusion.}
Proposals from the two dates are combined into one set of candidate regions, $\mathbf{M} = \mathcal{F}(\mathbf{M}_0, \mathbf{M}_1)$. A pair of instances from the two dates with IoU above 0.5 is treated as the same, persistent object and both are discarded; all other instances of either date are kept. Objects that appear, disappear or change footprint are therefore examined.

\subsection{Change Detection Module}
Methods that decide change by mask overlap or instance correspondence propagate every segmentation error into the change map.
\spotdiff instead uses semantic consistency as the decision signal and the proposals only to define \emph{where} it is evaluated.
For every candidate instance $m \in \mathbf{M}$ we compute its mean similarity
\begin{equation}
  s_m = \frac{1}{|m|} \sum_{(i,j) \in m} S(i,j),
\end{equation}
where $|m|$ is the number of pixels in the instance, and declare the instance changed if $s_m < \tau$.
The change mask is the union of all changed instances,
\begin{equation}
  \mathbf{Y} = \bigvee_{m \in \mathbf{M}:\; s_m < \tau} m,
  \label{eq:decision}
\end{equation}
where $\vee$ is the pixel-wise logical OR.
Averaging over an instance makes the decision robust to isolated noisy pixels in $\mathbf{S}$, and a semantic disagreement counts as change only when it is supported by an object proposal, which suppresses changes of other categories such as seasonal vegetation when detecting buildings.
We use a single threshold for all datasets ($\tau = 0.6$, \cref{sec:ablation}).

\paragraph{Class-agnostic variant.}
Without a prompt, $\mathcal{R}$ is skipped and change is decided per pixel, $\mathbf{Y} = \mathbf{1}[\mathbf{S} < \tau]$, which detects any semantic change.
We use this variant for the cross-resolution examples in \cref{fig:xres}.

\subsection{Plug-and-play design}
The Similarity Module is the composition $\mathcal{S} = \mathrm{cos} \circ (g_\phi \circ f_\theta)$, the RoI module is any promptable segmenter, and the Change Detection Module depends on neither.
A stronger encoder, upsampler or segmenter can therefore be substituted without modifying or retraining the rest of the pipeline.
In this paper, swapping the DINOv3 backbone from ViT-L to ViT-7B requires no change other than the checkpoint.

\begin{table}[t]
\centering
\small
\setlength{\tabcolsep}{4pt}
\begin{tabular}{@{}lcccc@{}}
\toprule
Method & Params (M) & TFLOPs & Time (s) & F1 \\
\midrule
OmniOVCD~\cite{zhang2026omniovcd} & 840 & -- & -- & 80.40 \\
AdaptOVCD~\cite{dou2026adaptovcd} & 1,365 & -- & -- & 68.00 \\
\spotdiff (ViT-L) & 1,144 & 134.5 & 9.6 & 80.10 \\
\spotdiff (ViT-7B) & 7,557 & 311.4 & $\approx$25 & \textbf{82.46} \\
\bottomrule
\end{tabular}
\caption{\textbf{Compute} for one $1024\times1024$ pair; F1 on LEVIR-CD (\%). Params are frozen foundation-model weights (none are trained). FLOPs counted with PyTorch's \texttt{FlopCounterMode} (1 multiply-add = 2 FLOPs). \spotdiff: SAM~3 840M, AnyUp 0.9M, DINOv3 303M (ViT-L) or 6,716M (ViT-7B); TFLOPs per pair = encoder 8.3 / 132.2, AnyUp 24.3 / 77.3, SWIMS 101.9 (91.7 of it at $2\times$). OmniOVCD: SAM~3 weights; AdaptOVCD: sum of the parameter counts in its paper. Time on one H100. ``--'': not reported.}
\label{tab:compute}
\end{table}

\paragraph{Cost.}
\cref{tab:compute} lists parameters and FLOPs. For a 1024$\times$1024 pair on one NVIDIA H100 GPU, our unoptimized implementation takes 9.6\,s with ViT-L: 4.1\,s for the similarity map, 5.4\,s for SWIMS on both dates and 0.1\,s for the decision. With ViT-7B the similarity map takes 19.8\,s (about 25\,s in total). The class-agnostic variant needs only the similarity map.

%% file: sections/4_experiments.tex
\section{Experiments}
\label{sec:exp}

\subsection{Setup}

\paragraph{Datasets.}
\textbf{LEVIR-CD}~\cite{chen2020levir} contains 637 pairs of 1024$\times$1024 images at 0.5\,m from 20 regions in Texas, with building construction and demolition over 5--14 years; we use the standard test split.
\textbf{WHU-CD}~\cite{ji2018whu} covers building change over a 20.5\,km$^2$ urban area in aerial imagery; as the $256\times256$ patch split of~\cite{zhang2026omniovcd} is not public, we evaluate on 154 $1024\times1024$ test tiles, so WHU-CD numbers are not strictly comparable with the baselines.
\textbf{S2Looking}~\cite{shen2021s2looking} contains 5,000 pairs from several satellites with side-looking (off-nadir) views of rural building change.
\textbf{SECOND}~\cite{yang2021second} contains 4,662 semantic change pairs of 512$\times$512 pixels with six land-cover classes; following~\cite{li2025dynamicearth,zhang2026omniovcd}, each class is evaluated as binary change on the 1,694-pair test split.
For qualitative cross-resolution results we use PlanetScope imagery at 3\,m from DynamicEarthNet~\cite{toker2022dynamicearthnet} and Sentinel-2 RGB imagery at 10\,m from OSCD~\cite{daudt2018oscd}.

\paragraph{Metrics.}
We report IoU, F1, precision and recall of the change class, in \%, with F1 as the primary metric.

\paragraph{Baselines.}
We compare with unsupervised methods~\cite{celik2009pcakm,elamin2016cnncd,du2019dsfa,saha2019dcva,tang2021gmcd,bovolo2006cva}, the training-free AnyChange~\cite{zheng2024anychange} and SCM~\cite{tan2024scm}, the strongest DynamicEarth pipelines~\cite{li2025dynamicearth}, OmniOVCD~\cite{zhang2026omniovcd}, AdaptOVCD~\cite{dou2026adaptovcd} and OpenDPR~\cite{guo2026opendpr}.
This work was carried out in March 2026, so we compare with training-free methods published before April 2026; later work is not included. Baseline numbers are taken from~\cite{zhang2026omniovcd}, which re-implements the DynamicEarth pipelines under identical settings; AdaptOVCD and OpenDPR numbers are taken from their papers.

\paragraph{Implementation.}
We use DINOv3~\cite{simeoni2025dinov3} ViT-7B/16 (default) or ViT-L/16 (300M parameters) as the encoder, the released AnyUp~\cite{wimmer2025anyup} model as the upsampler, and SAM~3~\cite{carion2025sam3} with text prompts for region proposals.
Unless noted, SWIMS uses scales $\{1\times, 2\times\}$, $\tau = 0.6$, and building change is prompted with ``roof'', the best prompt in \cref{tab:prompt}. On SECOND, each class is prompted with one or more class names and the instances of all its prompts are pooled: building (``building''), tree (``tree'', ``forest''), water (``water'', ``river'', ``pond'', ``sea''), low vegetation (``low vegetation'', ``grass''), surface (``the ground'', ``bare land'') and playground (``football court'', ``basketball court'', ``baseball court''). An instance is discarded if more than half of it is covered by negative prompts (``background''; for surface also the other classes, ``cropland'' and ``sports field''; for playground also ``building'' and ``blue roof''), and the covered pixels are removed otherwise.
No training is performed; all experiments run on a single NVIDIA H100 (80\,GB) or H200 (141\,GB) GPU.

\begin{table*}[t]
\centering
\small
\setlength{\tabcolsep}{6pt}
\begin{tabular}{@{}lcccccc@{}}
\toprule
\multirow{2}{*}{Method} & \multicolumn{2}{c}{LEVIR-CD} & \multicolumn{2}{c}{WHU-CD} & \multicolumn{2}{c}{S2Looking} \\
\cmidrule(lr){2-3}\cmidrule(lr){4-5}\cmidrule(l){6-7}
 & IoU & F1 & IoU & F1 & IoU & F1 \\
\midrule
PCA-KM~\cite{celik2009pcakm} & 4.8 & 9.1 & 5.4 & 10.2 & -- & -- \\
CNN-CD~\cite{elamin2016cnncd} & 7.0 & 13.1 & 4.9 & 9.4 & -- & -- \\
DSFA~\cite{du2019dsfa} & 4.3 & 8.2 & 4.1 & 7.8 & -- & -- \\
DCVA~\cite{saha2019dcva} & 7.6 & 14.1 & 10.9 & 19.6 & -- & -- \\
GMCD~\cite{tang2021gmcd} & 6.1 & 11.6 & 10.9 & 19.7 & -- & -- \\
CVA~\cite{bovolo2006cva} & -- & 12.2 & -- & -- & -- & 5.8 \\
DINOv2+CVA~\cite{zheng2024anychange} & -- & 17.3 & -- & -- & -- & 4.3 \\
AnyChange-H~\cite{zheng2024anychange} & -- & 23.0 & -- & -- & -- & 6.4 \\
SCM~\cite{tan2024scm} & 18.8 & 31.7 & 18.6 & 31.3 & -- & -- \\
SAM-DINO-SegEarth-OV$^*$~\cite{li2025dynamicearth} & 33.0 & 49.7 & 35.8 & 52.8 & 22.5 & 36.7 \\
SAM-DINOv2-SegEarth-OV$^*$~\cite{li2025dynamicearth} & 36.6 & 53.6 & 38.8 & 55.9 & 23.9 & 38.5 \\
SAM2-DINOv2-SegEarth-OV$^*$~\cite{li2025dynamicearth} & 33.8 & 50.5 & 36.8 & 53.8 & 23.1 & 37.6 \\
APE-/-DINO$^*$~\cite{li2025dynamicearth} & 53.5 & 69.7 & 54.5 & 70.5 & 10.1 & 18.4 \\
APE-/-DINOv2$^*$~\cite{li2025dynamicearth} & 50.0 & 66.7 & 55.2 & 71.1 & 5.3 & 10.1 \\
OmniOVCD~\cite{zhang2026omniovcd} & 67.2 & 80.4 & 66.5 & 79.9 & 24.5 & 39.4 \\
AdaptOVCD~\cite{dou2026adaptovcd} & 51.52 & 68.00 & 61.99 & 76.53 & -- & -- \\
OpenDPR~\cite{guo2026opendpr} & 44.8 & 61.9 & 54.3 & 70.4 & -- & -- \\
\midrule
\spotdiff (ViT-L) & 66.81 & 80.10 & 75.53$^\dagger$ & 86.06$^\dagger$ & 20.39 & 33.87 \\
\spotdiff (ViT-7B) & \textbf{70.15} & \textbf{82.46} & \textbf{76.16}$^\dagger$ & \textbf{86.47}$^\dagger$ & \textbf{28.66} & \textbf{44.55} \\
\bottomrule
\end{tabular}
\caption{\textbf{Training-free building change detection} (IoU / F1, \%). Baselines from~\cite{zhang2026omniovcd} except AdaptOVCD and OpenDPR (from their papers); $^*$re-implemented there under identical settings; ``--'': not reported. \spotdiff uses SWIMS $\{1\times,2\times\}$, $\tau=0.6$ and the prompt ``roof''. $^\dagger$Evaluated on $1024\times1024$ test tiles rather than the $256\times256$ patches of~\cite{zhang2026omniovcd}.}
\label{tab:building}
\end{table*}

\begin{table*}[t]
\centering
\small
\setlength{\tabcolsep}{2.9pt}
\begin{tabular}{@{}lcccccccccccccc@{}}
\toprule
\multirow{2}{*}{Method} & \multicolumn{2}{c}{Building} & \multicolumn{2}{c}{Tree} & \multicolumn{2}{c}{Water} & \multicolumn{2}{c}{Low veg.} & \multicolumn{2}{c}{Surface} & \multicolumn{2}{c}{Playground} & \multicolumn{2}{c}{Class avg.} \\
\cmidrule(lr){2-3}\cmidrule(lr){4-5}\cmidrule(lr){6-7}\cmidrule(lr){8-9}\cmidrule(lr){10-11}\cmidrule(lr){12-13}\cmidrule(l){14-15}
 & IoU & F1 & IoU & F1 & IoU & F1 & IoU & F1 & IoU & F1 & IoU & F1 & IoU & F1 \\
\midrule
SAM-DINOv2-SegEarth-OV$^*$~\cite{li2025dynamicearth} & 38.8 & 56.0 & 15.6 & 27.0 & 15.3 & 26.5 & 21.3 & 35.1 & 26.7 & 42.1 & 22.6 & 36.8 & 23.4 & 37.3 \\
SAM2-DINOv2-SegEarth-OV$^*$~\cite{li2025dynamicearth} & 36.3 & 53.3 & 15.8 & 27.3 & 14.6 & 25.5 & 20.1 & 33.4 & 19.6 & 32.7 & 24.6 & 39.5 & 21.8 & 35.3 \\
APE-/-DINO$^*$~\cite{li2025dynamicearth} & 29.1 & 45.1 & 9.7 & 17.7 & 12.3 & 22.0 & -- & -- & -- & -- & 25.6 & 40.8 & 12.8 & 21.0 \\
APE-/-DINOv2$^*$~\cite{li2025dynamicearth} & 31.9 & 48.3 & 10.6 & 19.2 & 12.2 & 21.7 & -- & -- & -- & -- & 25.0 & 40.0 & 13.3 & 21.5 \\
OmniOVCD~\cite{zhang2026omniovcd} & 45.2 & 62.3 & 16.7 & 28.6 & 21.2 & 35.0 & 24.5 & 39.3 & 27.7 & 43.4 & 27.0 & 42.4 & 27.1 & 41.8 \\
AdaptOVCD~\cite{dou2026adaptovcd} & 46.85 & 63.81 & 12.67 & 22.49 & 23.34 & 37.84 & 22.15 & 36.27 & \textbf{33.99} & \textbf{50.74} & 29.28 & 45.30 & 28.05 & 42.74 \\
OpenDPR~\cite{guo2026opendpr} & 42.4 & 59.5 & \textbf{20.9} & \textbf{34.5} & 17.5 & 29.8 & 23.0 & 37.4 & 30.2 & 46.4 & \textbf{37.3} & \textbf{54.3} & 28.6 & 43.7 \\
\midrule
\spotdiff (ViT-7B) & \textbf{55.53} & \textbf{71.41} & 16.25 & 27.96 & \textbf{27.60} & \textbf{43.26} & \textbf{28.69} & \textbf{44.59} & 23.17 & 37.63 & 27.55 & 43.20 & \textbf{29.34} & \textbf{44.23} \\
\bottomrule
\end{tabular}
\caption{\textbf{Per-class change detection on SECOND} (IoU / F1, \%), each class evaluated as binary change following~\cite{li2025dynamicearth}. Baselines from~\cite{zhang2026omniovcd} except AdaptOVCD and OpenDPR (from their papers; OpenDPR in its multi-class inference setting); $^*$re-implemented there; ``--'' denotes a score close to 0; AdaptOVCD's class average is the mean of its six per-class values, computed by us. \spotdiff's class average is computed from one complete run of all six classes and differs slightly from the mean of the row (29.80 IoU) because of run-to-run variation in SAM~3 proposals.}
\label{tab:second}
\end{table*}

\subsection{Main results}

\paragraph{Building change.}
\cref{tab:building} compares \spotdiff with unsupervised and training-free methods.
On LEVIR-CD, \spotdiff with ViT-7B reaches 82.46 F1 and 70.15 IoU, 2.06 F1 and 2.95 IoU above OmniOVCD and 14.46 F1 above AdaptOVCD, with 74.70 precision and 92.00 recall.
With the smaller ViT-L encoder (300M parameters) it reaches 80.10 F1, on par with OmniOVCD.
For reference, fully supervised methods trained on the LEVIR-CD training set reach about 90 F1~\cite{ding2024samcd,codegoni2023tinycd}, so \spotdiff closes most of the gap without a single labelled pair.
On WHU-CD and S2Looking, \spotdiff reaches 86.47 and 44.55 F1 with ViT-7B, compared with 79.9 and 39.4 for OmniOVCD (WHU-CD on a different test tiling, see \cref{sec:exp}). With ViT-L, it reaches 86.06 F1 on WHU-CD but trails OmniOVCD on S2Looking (33.87 F1), where the off-nadir views benefit most from the larger encoder.

\begin{figure*}[!tp]
  \centering
  \includegraphics[width=\linewidth,height=0.8\textheight,keepaspectratio]{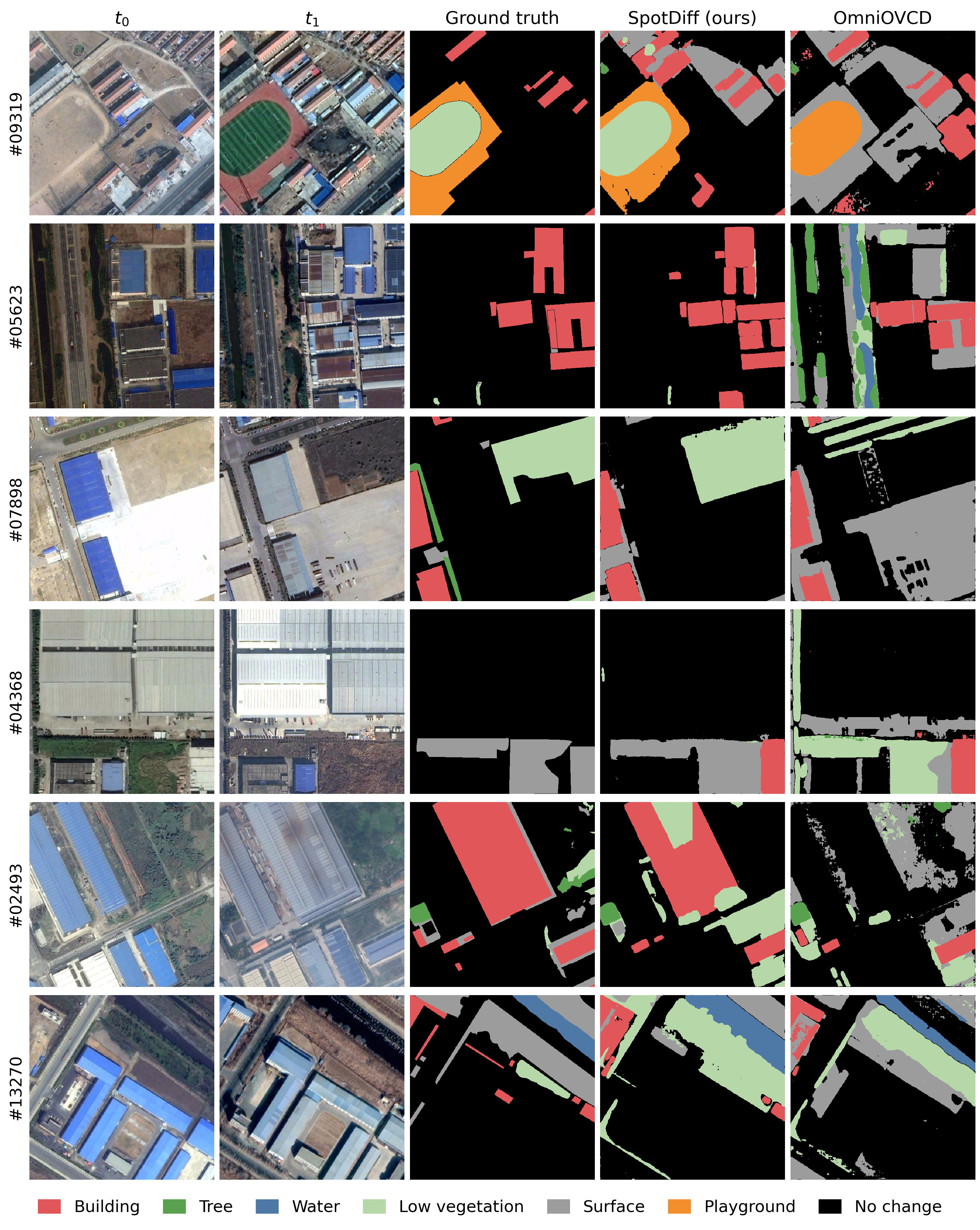}
  \caption{\textbf{Qualitative multi-class results on SECOND} (selected test pairs, image IDs on the left). Columns: image at $t_0$, image at $t_1$, ground truth, \spotdiff prediction, OmniOVCD prediction. Each class is detected as binary change with its own prompts; colours indicate the changed class (ground truth: class at $t_1$; black: no change). Where several classes fire on a pixel, \spotdiff shows the class of the instance with the lowest similarity; OmniOVCD exposes no instance score, so its overlaps follow a fixed priority (playground, water, building, tree, low vegetation, surface).}
  \label{fig:qual_second}
\end{figure*}

\paragraph{Multi-class change.}
\cref{tab:second} reports per-class results on SECOND.
\spotdiff improves over OmniOVCD on building (+10.3 IoU), water (+6.4), low vegetation (+4.2) and playground (+0.6), and is slightly lower on tree ($-$0.5) and surface ($-$4.5). Against AdaptOVCD and OpenDPR, it is best on building, water and low vegetation, while OpenDPR is stronger on tree and playground and AdaptOVCD on surface.
The gains concentrate on classes with clear object structure, which is where instance-level comparison of semantic features helps most; amorphous classes such as impervious surface are harder to delineate as instances, which limits the region prior.
Averaged over all six classes, \spotdiff reaches 29.34 IoU and 44.23 F1, against 27.1 and 41.8 for OmniOVCD, 28.05 and 42.74 for AdaptOVCD, and 28.6 and 43.7 for OpenDPR.
\cref{fig:qual_second} shows selected test pairs covering all six classes. \spotdiff recovers new playgrounds, buildings and vegetation patches with object-aligned boundaries, while OmniOVCD often marks large unchanged areas as changed surface (\#09319, \#05623, \#07898). Errors remain on mixed scenes: in \#02493 part of a rebuilt roof is labelled low vegetation, and small tree changes are partly missed.

\begin{figure*}[t]
  \centering
  \includegraphics[width=\linewidth]{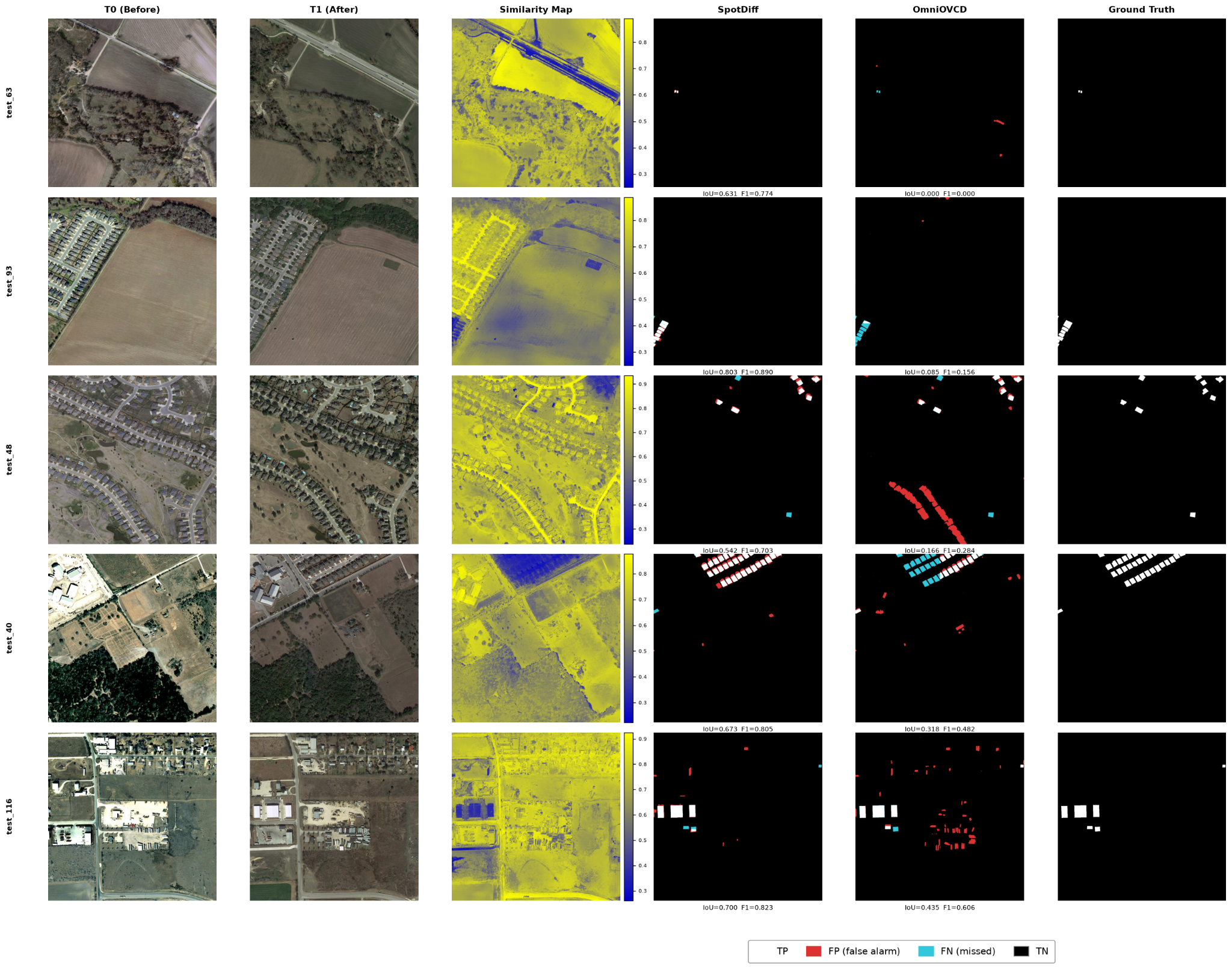}
  \caption{\textbf{Qualitative comparison on LEVIR-CD.} Columns: image at $t_0$, image at $t_1$, \spotdiff similarity map, \spotdiff prediction, OmniOVCD prediction, ground truth. White: true positive; red: false positive; cyan: false negative.}
  \label{fig:qual}
\end{figure*}

\paragraph{Qualitative comparison.}
\cref{fig:qual} compares \spotdiff with OmniOVCD on LEVIR-CD.
OmniOVCD produces false alarms along roads and on bare ground and misses several small buildings, while \spotdiff keeps its predictions on changed buildings.
The similarity maps show why: unchanged roofs and roads stay highly similar across dates, while new buildings appear as compact low-similarity regions.

\begin{figure*}[t]
  \centering
  \includegraphics[width=\linewidth]{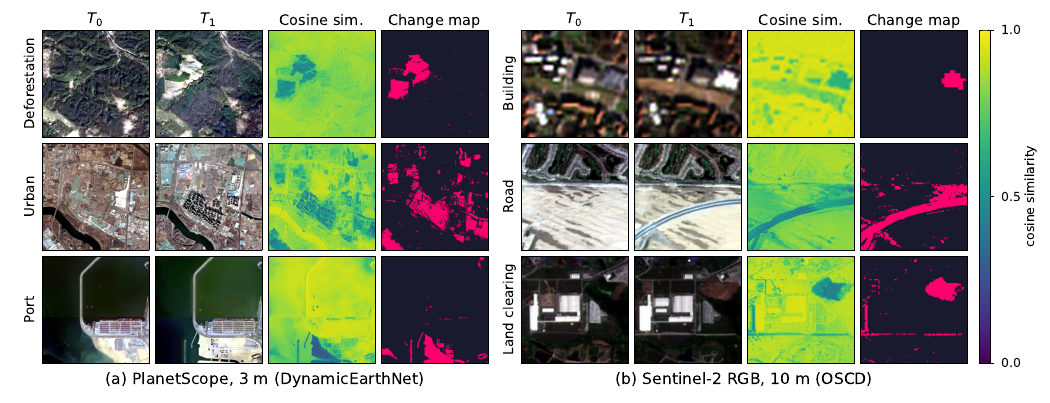}
  \caption{\textbf{Cross-resolution results} of the class-agnostic variant ($\tau = 0.7$) on medium-resolution imagery, using the same models and settings as on sub-meter data. Magenta: pixels with cosine similarity below $\tau$ (changed). (a) PlanetScope at 3\,m: deforestation, urban and port changes. (b) Sentinel-2 RGB at 10\,m: building, road and land-clearing changes.}
  \label{fig:xres}
\end{figure*}

\paragraph{Cross-resolution transfer.}
\cref{fig:xres} applies the same pipeline, unchanged, to PlanetScope (3\,m) and Sentinel-2 (10\,m) imagery.
Deforestation, urban growth, new roads and land clearing appear as coherent low-similarity regions, and the change maps follow their boundaries despite the coarser ground sampling distance.
At 10\,m, isolated buildings span only a few pixels and are detected as small blobs, which marks the practical lower limit of the approach.
These results are qualitative, as neither dataset provides category-specific labels matching our prompts.

\subsection{Ablation studies}
\label{sec:ablation}

\begin{table}[t]
\centering
\small
\begin{tabular}{@{}lcccc@{}}
\toprule
Upsampler & F1 & IoU & Prec. & Rec. \\
\midrule
Nearest neighbor & 66.88 & 54.72 & 57.79 & 90.44 \\
Bilinear & 68.69 & 56.19 & 55.20 & 90.94 \\
AnyUp & \textbf{82.46} & \textbf{70.15} & \textbf{74.70} & \textbf{92.00} \\
\bottomrule
\end{tabular}
\caption{\textbf{Upsampler ablation} on LEVIR-CD (\%).}
\label{tab:upsampler}
\end{table}

\paragraph{Feature upsampling.}
\cref{tab:upsampler} replaces AnyUp with nearest-neighbor or bilinear interpolation of the DINOv3 features.
AnyUp gives the highest F1 and precision, 13.8 F1 above the bilinear interpolation used by prior feature-comparison pipelines~\cite{dou2026adaptovcd}: interpolation blurs features across object boundaries, so the similarity of small buildings is mixed with that of their surroundings, whereas image-guided upsampling keeps boundaries sharp.

\begin{table}[t]
\centering
\small
\setlength{\tabcolsep}{4pt}
\begin{tabular}{@{}llccccc@{}}
\toprule
Backbone & Scales & 0.5 & 0.6 & 0.65 & 0.7 & 0.8 \\
\midrule
\multirow{3}{*}{ViT-7B} & $1\times$ & 55.97 & 61.09 & 61.16 & 59.43 & 55.05 \\
 & $1\times$+$2\times$ & 78.54 & \textbf{82.46} & 82.11 & 80.12 & 76.42 \\
 & $1\times$+$2\times$+$4\times$ & 76.29 & 77.36 & 76.63 & 74.12 & 69.53 \\
\midrule
\multirow{3}{*}{ViT-L} & $1\times$ & 52.19 & 58.16 & 59.75 & 59.95 & 56.87 \\
 & $1\times$+$2\times$ & 73.64 & 80.10 & 81.22 & \textbf{81.29} & 77.91 \\
 & $1\times$+$2\times$+$4\times$ & 70.51 & 75.29 & 75.90 & 75.45 & 71.33 \\
\bottomrule
\end{tabular}
\caption{\textbf{SWIMS scales and threshold $\tau$} on LEVIR-CD (F1, \%). Columns are values of $\tau$.}
\label{tab:swims}
\end{table}

\paragraph{SWIMS and threshold.}
\cref{tab:swims} varies the SWIMS scales and $\tau$ for both encoders.
Adding the $2\times$ scale is the single largest effect in the pipeline: +21.4 F1 for ViT-7B (61.09 to 82.46 at $\tau = 0.6$) and +21.3 for ViT-L (59.95 to 81.29 at $\tau = 0.7$).
Adding $4\times$ lowers F1 for both encoders, because over-magnified windows contain uniform texture that SAM~3 segments into spurious instances.
With $\{1\times,2\times\}$, F1 stays within 0.4 of its maximum over $\tau \in [0.6, 0.65]$ for ViT-7B and $[0.65, 0.7]$ for ViT-L, so the method is not sensitive to the exact threshold.

\begin{figure}[t]
  \centering
  \includegraphics[width=\linewidth]{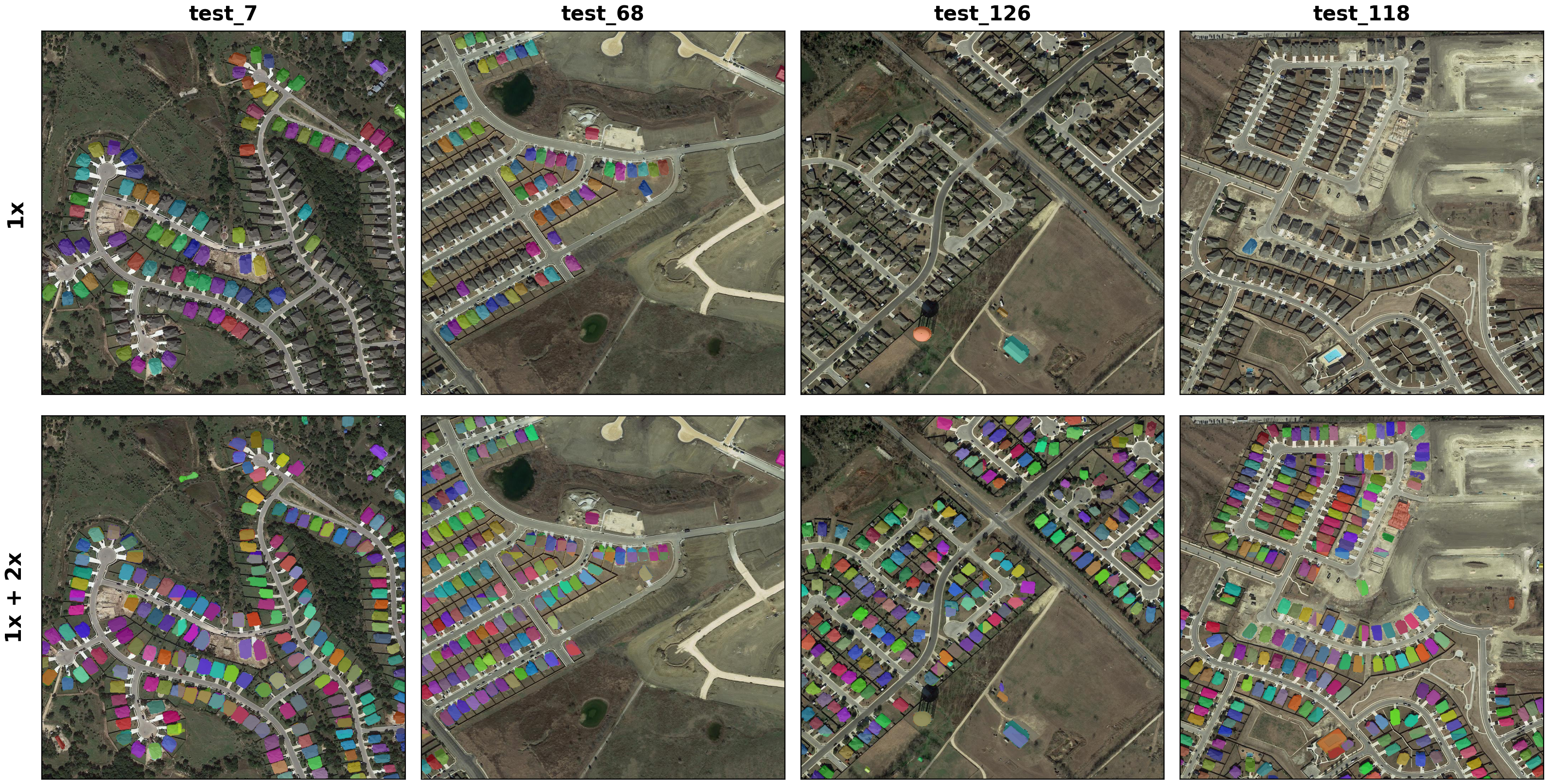}
  \caption{\textbf{SWIMS vs.\ single-scale inference} on LEVIR-CD. SAM~3 instances at $1\times$ (top) and with SWIMS $\{1\times,2\times\}$ (bottom).}
  \label{fig:swims}
\end{figure}

\cref{fig:swims} illustrates the effect: at a single scale SAM~3 misses most small residential buildings in a 1024$\times$1024 tile, and the $2\times$ windows recover them.

\begin{figure*}[t]
  \centering
  \includegraphics[width=\linewidth]{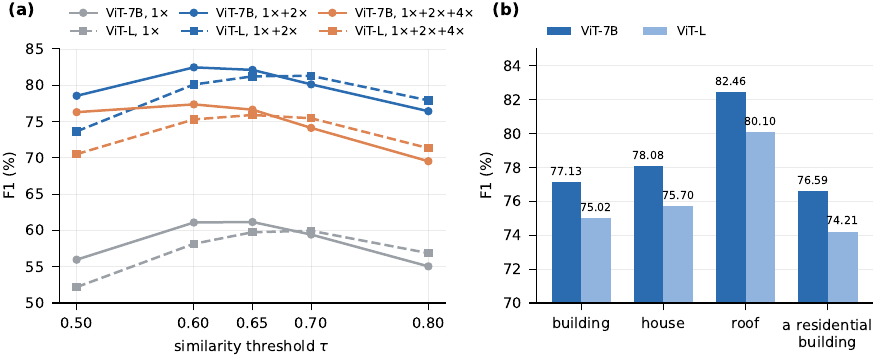}
  \caption{\textbf{Sensitivity on LEVIR-CD.} (a) F1 against the similarity threshold $\tau$ for each SWIMS scale setting and encoder (\cref{tab:swims}). (b) F1 for four synonymous SAM~3 prompts (\cref{tab:prompt}).}
  \label{fig:sensitivity}
\end{figure*}

\cref{fig:sensitivity}(a) shows the same trends: the $2\times$ scale lifts every curve, and all curves peak in $\tau \in [0.6, 0.7]$, so a single fixed threshold works for both encoders.

\begin{table}[t]
\centering
\footnotesize
\setlength{\tabcolsep}{2.6pt}
\begin{tabular}{@{}llccccc@{}}
\toprule
Backbone & Prompt & Prec. & Rec. & F1 & IoU & Acc. \\
\midrule
\multirow{4}{*}{ViT-7B} & building & 69.95 & 85.96 & 77.13 & 62.78 & 97.40 \\
 & house & 69.83 & 88.54 & 78.08 & 64.04 & 97.47 \\
 & roof & \textbf{74.70} & \textbf{92.00} & \textbf{82.46} & \textbf{70.15} & \textbf{98.01} \\
 & a residential building & 72.23 & 81.51 & 76.59 & 62.06 & 97.46 \\
\midrule
\multirow{4}{*}{ViT-L} & building & 70.77 & 79.81 & 75.02 & 60.03 & 97.29 \\
 & house & 70.23 & 82.11 & 75.70 & 60.91 & 97.32 \\
 & roof & \textbf{75.29} & \textbf{85.57} & \textbf{80.10} & \textbf{66.81} & \textbf{97.83} \\
 & a residential building & 72.92 & 75.54 & 74.21 & 58.99 & 97.32 \\
\bottomrule
\end{tabular}
\caption{\textbf{Text-prompt sensitivity} on LEVIR-CD (\%, SWIMS $\{1\times,2\times\}$, $\tau=0.6$).}
\label{tab:prompt}
\end{table}

\paragraph{Text prompt.}
\cref{tab:prompt} and \cref{fig:sensitivity}(b) vary the SAM~3 prompt for building change.
``Roof'' is best for both encoders, 5.3 (ViT-7B) and 5.1 (ViT-L) F1 above ``building'', likely because roofs are what a building looks like from above.
Across the four synonymous prompts F1 varies by up to 5.9 points, so the prompt is a meaningful, training-free knob, and also a source of variance that users of text-prompted change detection should be aware of.

%% file: sections/5_conclusion.tex
\section{Conclusion}
\label{sec:conclusion}

We presented \spotdiff, a training-free, plug-and-play framework for zero-shot remote sensing change detection.
It composes a self-supervised encoder, a universal feature upsampler and a promptable segmentation model, and decides change by comparing pixel-resolution features inside text-guided region priors made reliable by multi-scale inference.
Without change labels or fine-tuning, \spotdiff reaches 82.46 F1 on LEVIR-CD, improves building change detection on SECOND by 10.3 IoU over the strongest prior training-free method, and transfers to 3\,m and 10\,m imagery unchanged.
Because every component is replaceable, \spotdiff should improve as foundation models do.

\paragraph{Limitations.}
The similarity threshold is fixed; adaptive thresholding from image statistics could improve robustness.
The method assumes co-registered images, inference is dominated by multi-scale SAM~3 and, with ViT-7B, by the encoder, and accuracy depends on the text prompt (\cref{tab:prompt}).
Region priors help most for object-like categories and less for amorphous ones such as impervious surface.
\spotdiff is evaluated on RGB optical imagery only; multispectral and SAR inputs are left to future work.

\paragraph{Acknowledgements and LLM usage.}
Large language model (LLM) tools were used to assist with language editing, restructuring of the manuscript, \LaTeX{} formatting and preparation of some figures from our experimental results. All methods, experiments and results were designed, run and verified by the authors, who take full responsibility for the content of this paper.